\documentclass[letterpaper, 10 pt, conference]{ieeeconf}  

\IEEEoverridecommandlockouts                              

\usepackage{color,xcolor}
\usepackage{graphicx} 
\usepackage{booktabs}
\usepackage{multirow} 
\usepackage{threeparttable}
\usepackage{colortbl}
\usepackage{makecell}
\usepackage{amsmath} 
\usepackage{amssymb}  
\usepackage[normalem]{ulem}
\usepackage{hyperref} 
\usepackage{subcaption}

\definecolor{b}{HTML}{00B0F0}
\definecolor{r}{HTML}{DF0000}
\definecolor{c2}{HTML}{FBD9BD}

\makeatletter
\renewcommand{\maketag@@@}[1]{\hbox{\m@th\normalsize\normalfont#1}}%
\makeatother

\title{\LARGE \bf
A Vision-Language-Action Model for Adaptive Ultrasound-Guided Needle Insertion and Needle Tracking
\thanks{Accepted by ICRA 2026.}
\vspace{-2mm}
}

\author{
Yuelin Zhang, Qingpeng Ding, Longxiang Tang, Chengyu Fang, Shing Shin Cheng$^{*}$ 
\vspace{-2mm}
\thanks{Research reported in this work was supported in part by Research Grants Council (RGC) of Hong Kong (CUHK 14217822, CUHK 14207823, CUHK 14211425, T45-401/22-N, and AoE/E-407/24-N) and in part by Innovation and Technology Commission of Hong Kong (MHP/096/22, ITS/235/22, ITS/224/23, ITS/225/23, and Multi-scale Medical Robotics Center (InnoHK initiative)).}
\thanks{Yuelin Zhang and Qingpeng Ding are with the Department of Mechanical and Automation Engineering, The Chinese University of Hong Kong, Hong Kong.}
\thanks{Longxiang Tang is with the Department of Computer Science and Engineering, The Hong Kong University of Science and Technology, Hong Kong. \looseness=-1}
\thanks{Chengyu Fang is with the Shenzhen International Graduate School, Tsinghua University, China.}
\thanks{Shing Shin Cheng is with the Department of Mechanical and Automation Engineering, T Stone Robotics Institute, Shun Hing Institute of Advanced Engineering, Multi-Scale Medical Robotics Center, and Institute of Medical Intelligence and XR, The Chinese University of Hong Kong, Hong Kong.{\tt\small $^{*}$sscheng@cuhk.edu.hk}}%
}

\begin{document}

\maketitle
\thispagestyle{empty}
\pagestyle{empty}

\begin{abstract}

Ultrasound (US)-guided needle insertion is a critical yet challenging procedure due to dynamic imaging conditions and difficulties in needle visualization. 
Many methods have been proposed for automated needle insertion, but they often rely on hand-crafted pipelines with modular controllers, whose performance degrades in challenging cases.
In this paper, a Vision-Language-Action (VLA) model is proposed for adaptive and automated US-guided needle insertion and tracking on a robotic ultrasound (RUS) system. 
This framework provides a unified approach to needle tracking and needle insertion control, enabling real-time, dynamically adaptive adjustment of insertion based on the obtained needle position and environment awareness.
To achieve real-time and end-to-end tracking, a Cross-Depth Fusion (CDF) tracking head is proposed, integrating shallow positional and deep semantic features from the large-scale vision backbone.
To adapt the pretrained vision backbone for tracking tasks, a Tracking-Conditioning (TraCon) register is introduced for parameter-efficient feature conditioning.
After needle tracking, an uncertainty-aware control policy and an asynchronous VLA pipeline are presented for adaptive needle insertion control, ensuring timely decision-making for improved safety and outcomes.
Extensive experiments on both needle tracking and insertion show that our method consistently outperforms state-of-the-art trackers and manual operation, achieving higher tracking accuracy, improved insertion success rates, and reduced procedure time, highlighting promising directions for RUS-based intelligent intervention.

\end{abstract}

\section{INTRODUCTION}
Ultrasound (US)-guided needle insertion is frequently used in a variety of percutaneous procedures, including minimally invasive interventions like tissue biopsy, tumor ablation, and regional anesthesia \cite{liu2019deep}. As a non-invasive, portable, safe, and cost-effective imaging technique~\cite{richardson2017imaging}, US offers real-time intraoperative visualization of both the needle and surrounding tissue, thus helping to reduce the risk of inadvertent injuries.

Some US needle trackers have been proposed to provide real-time needle position feedback. 
However, even with accurate needle position feedback, precise needle insertion still requires skillful operators, which can be challenging in resource-limited settings. Robotic ultrasound (RUS) has therefore emerged as a transformative technology for automated needle insertion, offering enhanced precision and consistency.
Recent advancements have enabled RUS systems to perform complex tasks such as probe manipulation, optimal image acquisition, and needle steering, thereby reducing operator dependency and improving procedure outcomes \cite{chatelain20153d}.

There are many challenges during needle insertion, where procedure outcomes can be affected by occlusion, imaging artifacts, and intermittent needle invisibility \cite{kimbowa2024advancements}.
Traditional hand-crafted automated needle-insertion pipelines \cite{lapouge2020towards,chatelain20153d,khadem2016ultrasound} remain fragile when facing these challenges, which require not only \textbf{accurate needle tracking} for closed-loop control but also \textbf{generalizable context awareness} to proactively adapt to dynamic environments.
This highlights the need for better generalizability and high-level reasoning beyond conventional feature engineering methods.

Building upon the developments of Large-Language Models (LLMs) \cite{achiam2023gpt} and Vision-Language Models (VLMs) \cite{bai2025qwen2}, Vision-Language-Action (VLA) \cite{ma2024survey} models have gained increasing attention in both open-world tasks and medical fields, as they can both understand multimodal information and generate actionable outputs. 
Although VLA models demonstrate robust logical reasoning abilities and strong generalization to dynamic environments, research on VLA-based automated needle insertion and tracking remains limited due to their high computational and data requirements. Achieving satisfactory accuracy while maintaining efficiency in VLA models remains challenging.

In this work, a novel VLA framework for automated and adaptive US-guided needle insertion on a RUS system is proposed for the first time. 
Without reliance on external sensors or pre-registration, the proposed framework holds promise for safer, more standardized, and operator-independent clinical workflows than conventional visual servoing needle steering pipelines.
To enable adaptive needle insertion, accurate real-time needle position feedback is indispensable.
Instead of incorporating a separate tracker like traditional methods, a Cross-Depth Fusion (CDF) Tracking Head is introduced as an end-to-end tracking head to incorporate cross-layer features from the pretrained vision backbone.
Compared to hand-crafted non-end-to-end pipelines with independent needle trackers based on particle filter \cite{chatelain20153d} or one-stream pipelines with computationally expensive VLM-based object grounding \cite{ng2025endovla}, the proposed framework not only ensures end-to-end training but also maintains efficiency.
By further integrating the proposed learnable Tracking-Conditioning (TraCon) Register, the pretrained vision backbone is parameter-efficiently conditioned for robust tracking.

Based on the needle position feedback, the VLA then controls the needle to reach the target with the proposed uncertainty-aware control policy, which ensures procedural success and safety when facing uncertainty due to occlusion, imaging artifacts, or intermittent tip invisibility.
Leveraging large pretrained models, it provides better generalizability to dynamic US environments than traditional methods.
An asynchronous VLA pipeline is proposed to decouple visual analysis from action generation, which operates at different latencies, thereby achieving precise action generation while ensuring real-time tracking. 
Extensive experiments demonstrate the effectiveness of the proposed VLA-based tracking-control pipeline.

The contributions of our work are fourfold:
\begin{itemize}
    \item To provide accurate real-time needle position feedback for insertion control, a Cross-Depth Fusion (CDF) Tracking Head is developed to integrate shallow positional and deep semantic features, better adapting deep vision backbone for tracking tasks. 
    \item A Tracking-Conditioning (TraCon) Register is introduced as a lightweight learnable token for parameter-efficient transferable vision backbone adaptation, facilitating tracking-oriented conditioning.
    \item To achieve adaptive needle insertion, an uncertainty-aware control policy is implemented, ensuring procedural safety and success under imaging artifacts and procedural uncertainty.
    \item To accommodate different execution speeds of tracking head and LLM, an asynchronous VLA pipeline is proposed, simultaneously ensuring real-time needle tracking and precise insertion control.
\end{itemize}

\section{RELATED WORK}

\subsection{Ultrasound Needle Tracking and Robotic Ultrasound}
Convolutional neural networks (CNNs) and transformer-based deep learning approaches \cite{vaswani2017attention,zhang2024unified} have become widely adopted for tracking applications \cite{li2019siamrpn++,zhang2025motion}, including ultrasound (US) needle tracking \cite{mwikirize2019learning,zhang2025mambaxctrack,zhang2025mrtrack}.
Given that the needle tip can be obscured by noise and artifacts, subsequent works employ segmentation of the needle shaft as a precursor to tip localization \cite{wijata2024needle}.
Another strategy is to incorporate temporal needle motion information into tracking.
In \cite{yan2023learning}, a motion prediction mechanism is combined with a visual tracker to enable motion cues beyond vision information.
However, after acquiring needle position, the existing needle trackers failed to be integrated with control models to achieve intelligent intervention with needle position feedback.

To improve patient outcomes, the robotic ultrasound (RUS) systems are increasingly being integrated with artificial intelligence to enable autonomous procedures. 
Khadem et al. \cite{khadem2016ultrasound} introduced an autonomous US-guided intervention system that leverages nonlinear model predictive control (MPC) for accurate and adaptive needle targeting in soft tissue.
By integrating particle filter-based needle tracking with closed-loop visual servoing, a real-time 3D US-guided robotic needle steering system is proposed in \cite{chatelain20153d} for autonomous flexible needle insertion.
With further integration with segmentation-based state estimation, \cite{lapouge2020towards} proposed a US needle insertion system addressing noisy tissue features.
However, these modular feature-engineering pipelines with stacked separate modules remain brittle in dynamic environments, as they typically rely heavily on the accuracy of the physics-based model and require numerous hyperparameters for complex optimization, not to mention the impractical assumptions of persistent visibility and quasi-static status.
It thus motivates a VLA-based controller that unifies perception and action for context-aware needle insertion control.

\subsection{VLA Models in Medical Applications}
Based on LLMs and VLMs, Vision-Language-Action (VLA) models have rapidly evolved, enabling systems to perceive, interpret, and interact with complex environments through multimodal understanding and reasoning \cite{ma2024survey,wang2025trackvla}. 
In the medical domain, VLA models have shown great promises.
CapsDT \cite{he2025capsdt} is proposed as a VLA model for endoscopy capsule robot manipulation with diffusion transformer \cite{peebles2023scalable}, demonstrating state-of-the-art performance in complex gastrointestinal tasks.
More recently, EndoVLA \cite{ng2025endovla} introduces a dual-phase VLA framework designed for robotic endoscopy, enabling robust autonomous tracking of abnormal regions through end-to-end instruction imitating and reinforced with task-specific rewards. 
Although it achieves end-to-end object tracking and action generation within VLA, its use of a coupled structure for both tasks significantly compromises efficiency.

Due to the scarcity of RUS action datasets and the inherently high dynamics, noise, and indistinct target features of US images compared to other modalities, such VLA models for US procedures have yet to be proposed, despite their potential effectiveness.

\begin{figure}
    \centering
    \includegraphics[width=0.95\linewidth]{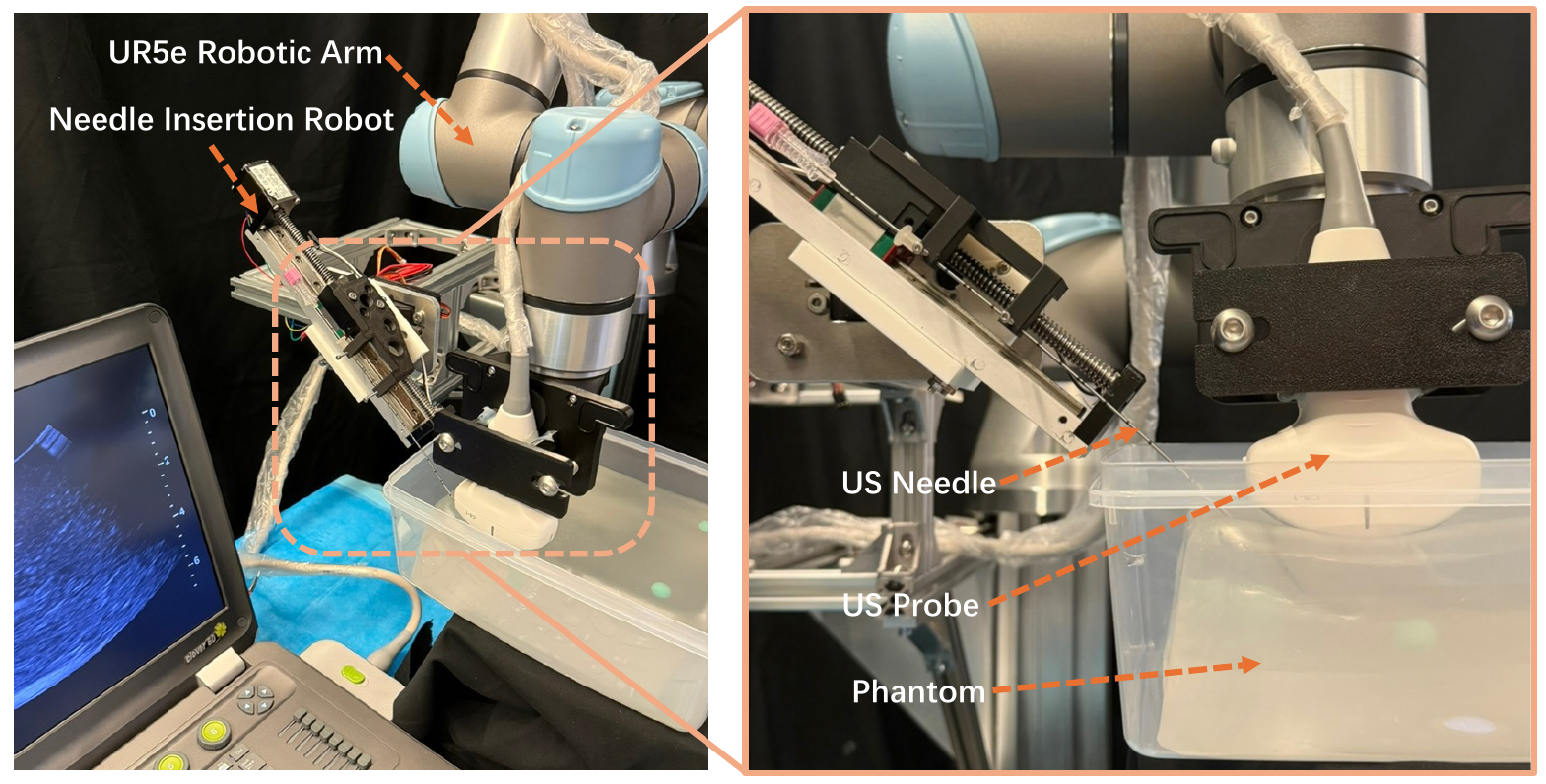}
    \caption{The proposed robotic ultrasound (RUS) platform.}
    \label{fig_rus}
\end{figure}

\begin{figure*}
    \centering
    \includegraphics[width=0.98\textwidth]{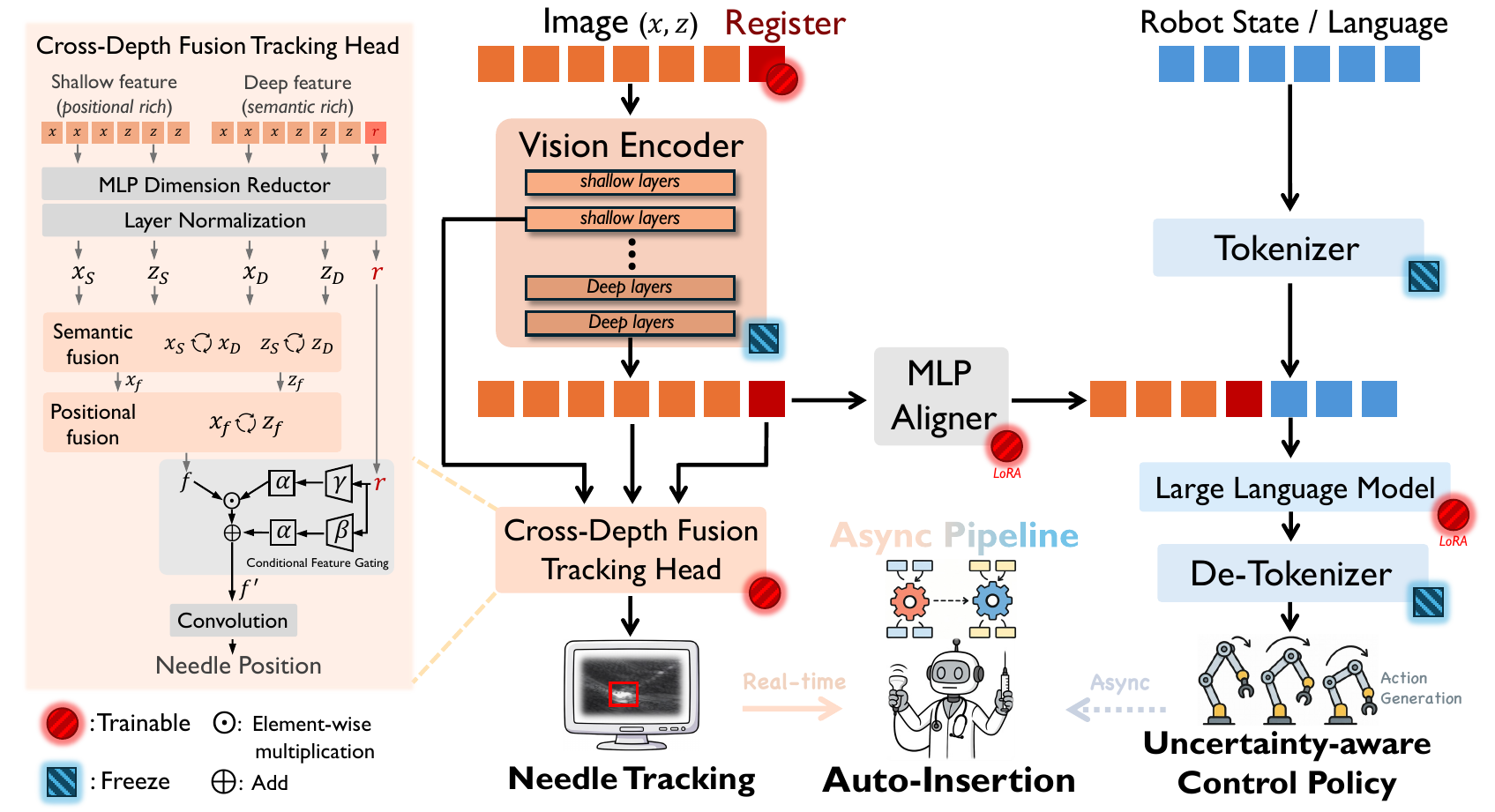}
    \caption{Structure overview. The proposed VLA framework integrates the CDF tracking head and the TraCon register with a large-scale vision encoder, and decouples tracking from the proposed uncertainty-aware control through an asynchronous pipeline. This design enables accurate real-time needle tracking and precise, context-aware insertion control.}
    \vspace{-5mm}
    \label{fig_overview}
\end{figure*}

\section{System Design}

\subsection{Robotic Ultrasound Platform}
A robotic ultrasound (RUS) platform was developed for automated needle insertion, which includes a needle insertion robot and a UR5e robotic arm, as shown in Fig.~\ref{fig_rus}.
The needle insertion robot consists of a linear manipulator for controlling the needle's axial displacement $x_{n}$ (defined in the needle coordinate frame) and a servo for adjusting the insertion angle $\theta_{n}$.
The UR5e robotic arm is for manipulating the US probe in three translational degrees of freedom $\textbf{x}_{p}=[x_p,y_p,z_p]$, expressed in the world coordinate frame.

\subsection{Vision-Language-Action Model Overview}
The proposed VLA is built based on a Qwen2.5-VL-3B model \cite{bai2025qwen2}.
As shown in Fig.~\ref{fig_overview}, the proposed VLA model adopts a pretrained vision encoder $\epsilon_V$ based on the Vision Transformer (ViT) \cite{dosovitskiy2020image}, a pretrained LLM $\phi_L$ for interpreting input and generating actions, and a separate Cross-Depth Fusion (CDF) tracking head $\phi_T$ specifically designed for real-time needle tracking. 
In addition, the proposed Tracking-Conditioning (TraCon) register is appended to input image embeddings as a learnable, parameter-efficient, and task-oriented token, which is then applied in tracking head and LLM for task adaptation.

In object tracking \cite{li2019siamrpn++}, the network typically performs prediction based on the search map $x$ and the template map $z$, where $z$ encodes the target object's appearance from the initial frame, while $x$ represents the region in subsequent frames where the target is likely to be located. The model identifies the object in $x$ that most closely matches $z$ and predicts a bounding box representing the target’s location.

The framework receives ultrasound imaging observation $O_t=[x_t,z_t]$ and the language instruction $I$, predicts needle position $P=\phi_T(\epsilon_V(O_t))$, and then predicts action $\mathcal{A}$ via an asynchronous inference pipeline to achieve automated adaptive needle insertion.

\subsection{Tracking-Conditioning Register}
The proposed Tracking-Conditioning (TraCon) register is a lightweight learnable token for task-oriented parameter-efficient fine-tuning (PEFT).
Most existing VLA models perform PEFT on the pretrained vision encoder by LoRA \cite{hu2022lora} to ensure that the pretrained model can be adapted to downstream tasks. 
However, such PEFT methods place high demands on the quantity and quality of labeled training data, while acquiring labeled datasets in medical scenarios is often difficult and expensive. 
Besides, in the proposed model, the output of the vision encoder is utilized by both the tracking head and the LLM. Directly fine-tuning the vision encoder would inevitably introduce performance degradation due to inconsistent training objectives. 

Instead of applying PEFT to the whole vision encoder, the proposed TraCon register $R\in \mathbb{R}^{B \times L_r \times C}$ ($L_r=4$ in this paper) is appended to the vision observation $O_t$ as a lightweight trainable token (0.5 M parameters), externally conditioning the frozen vision backbone for object tracking and aligning it with both the tracking head and the LLM.
After being encoded by the vision encoder $\epsilon_V$, the encoded TraCon register $\epsilon_V(R)$ is subsequently fed into both the CDF tracking head $\phi_T$ and LLM $\phi_L$. In CDF head, $\epsilon_V(R)$ is fused with vision embeddings via the proposed Conditional Feature Gating. In LLM, $\epsilon_V(R)$ is inserted into the input sequence to facilitate cross-modal knowledge transfer.

\begin{figure*}
    \centering
    \includegraphics[width=\textwidth]{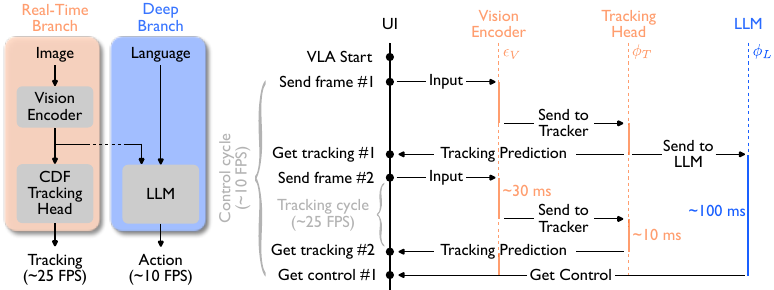}
    \caption{Asynchronous VLA pipeline (left) and its sequence diagram (right). The runtime of the tracking cycle and control cycle is measured while they are running simultaneously on a single NVIDIA A800 GPU. The lengths in the sequence diagram are for illustration only.}
    \label{fig_async}
\end{figure*}

\subsection{Cross-Depth Fusion Tracking Head}
The Cross-Depth Fusion (CDF) tracking head $\phi_T$ is proposed as a dedicated module for needle tracking, which is shown in Fig.~\ref{fig_overview}.
In CDF head, multi-depth features from the ViT vision encoder $\epsilon_V$ are simultaneously incorporated to learn shallow positional information and deep semantic information, since the shallow layer outputs in ViT predominantly encode positional information due to their proximity to the input embeddings and limited receptive field, whereas deeper layer outputs capture more semantic information as a result of accumulated global context and hierarchical feature abstraction \cite{dosovitskiy2020image}. \looseness=-1  

Specifically, shallow features and deep features (together with $\epsilon_V(R)$) will first go through a MLP dimension reductor (to reduce backbone dimension 2048 to internal dimension $C=512$) and a layer normalization \cite{ba2016layer} to get $x_S$, $z_S$, $x_D$, $z_D$, and $r$, respectively, where $x_S$ and $z_S$ are shallow outputs of search and template maps, $x_D$ and $z_D$ are deep outputs.
To incorporate deep semantic features into tracking prediction without compromising the shallow positional features, a cross-depth semantic fusion is first performed separately to $(x_S, x_D)$ and $(z_S, z_D)$ to get $x_f$ and $z_f$.
The core of semantic fusion can be formulated as:
\begin{equation}
SemFus = (\text{softmax}(\frac{Q_{ch} (K_{ch})^\text{T}}{\sqrt{L}}) V_{ch})^\text{T} \in \mathbb{R}^{B \times L \times C},
\end{equation}
where $Q_{ch}=\text{Linear}(q)^\text{T}$, $K_{ch}=\text{Linear}(k)^\text{T}$, and $V_{ch}=\text{Linear}(v)^\text{T}$ are channel sequences in the dimension of $B \times C \times L$.
The full semantic fusion block employs a standard transformer-style architecture \cite{vaswani2017attention}, which includes a residual shortcut, layer normalization \cite{ba2016layer}, and a MLP applied after the attention module.

With the cross-depth semantically enhanced maps $x_f$ and $z_f$, the next stage is the positional correlation. The positional correlation correlates $x_f$ with $z_f$ by modeling positional attention to precisely localize target $z_f$ on search map $x_f$:
\begin{equation}
PosCor = \text{softmax}(\frac{Q (K)^\text{T}}{\sqrt{C}}) V \in \mathbb{R}^{B \times L \times C},
\end{equation}
where $Q=\text{Linear}(q)$, $K=\text{Linear}(k)$, and $V=\text{Linear}(v)$ are all in $B \times L \times C$.
Similar to semantic fusion, the remaining parts of positional correlation also follow the standard transformer block design \cite{vaswani2017attention}.

Finally, before predicting the bounding box by a convolution layer ($stride=1$, $kernel=1 \times 1$), the fused map $f$ is integrated with register $r$ using the proposed Conditional Feature Gating. This operation conditions the primary vision feature map $f$ on auxiliary register embeddings that capture task-specific priors, enabling dynamic modulation without altering the spatial structure of $f$. It is formulated by:
\begin{equation}
f' = f \odot (\alpha \cdot \gamma(r)) + \alpha \cdot \beta(r),
\end{equation}
where $\gamma,\beta \in \mathbb{R}^{B \times C}$ are linear transformations, and $\alpha$ is a learnable scalar gate.
Compared with the existing FiLM (Feature-wise Linear Modulation) \cite{perez2018film}, a learnable scalar $\alpha$ is added to adaptively control the modulation strengths.
This mechanism enhances feature adaptability by conditioning on register-derived priors, promoting task-specific refinements in visual tracking. 

With this regressive CDF tracking head, the proposed VLA model possesses several intrinsic advantages on tracking. 
Existing VLA models with object tracking like EndoVLA \cite{ng2025endovla} typically couples action generation and tracking together to the LLM, where tracking is accomplished in an auto-regressive manner with LLM-based object grounding.
This results in inefficiency, with tracking and control sharing a low-speed pipeline running at 2 FPS, which falls significantly short of real-time requirements.
In contrast, by leveraging the dedicated CDF tracking head with decoupled action and tracking pipelines, the proposed VLA achieves significantly higher tracking \textbf{efficiency} ($\sim25$ FPS) with \textbf{fewer parameters}.
Furthermore, compared to hand-crafted models that combine separate needle trackers and insertion controllers \cite{khadem2016ultrasound,chatelain20153d,lapouge2020towards}, the CDF head and action generator (LLM) share the same pretrained vision backbone, enabling higher consistency between tracking and control.

\subsection{Asynchronous VLA Pipeline}
After obtaining the needle position, the next stage is to generate action with LLM for insertion control.
If tracking waits in a blocking manner for the LLM to complete before processing the next frame, like the existing one-stream methods \cite{ng2025endovla}, real-time tracking then cannot be guaranteed. To address this issue, an asynchronous VLA pipeline is proposed, as shown in Fig.~\ref{fig_async}.

The proposed pipeline includes a real-time branch for needle tracking ($\sim25$ FPS) and a deep branch for action generation ($\sim10$ FPS). 
After the vision encoder $\epsilon_V$ finishes encoding one frame, the tracking will then start. Once the needle tracking prediction $P$ is obtained by $\phi_T$, it will be sent to both user interface (UI) for display and LLM $\phi_L$ for action generation.  
While the LLM generates actions, the vision encoding and tracking prediction for the next frame start immediately. Once the LLM completes action prediction, it will receive the latest available vision embedding without waiting to initiate the next round of action generation.

\subsection{Uncertainty-Aware Control Policy}
In US-guided needle insertion, the tip visibility is dynamic and fragile. 
When the tip fades, before the change is detected and reacted upon, the needle will lose tracking and a steady insertion speed can translate into millimeters of unseen advancement, leading to \textbf{discontinuities} in tip visualization and \textbf{clinical danger} near critical structures.
Instead of relying on the operator to subjectively determine when to pause the needle advancement,
a VLA-based uncertainty-aware control policy is proposed, following the control law by skilled operators: \textit{see what you’re doing}, and act more slowly when uncertainty increases (uncertainty is defined as the inverse of tip visibility) \cite{sato2024basic,casanova2014effect}. 

After the CDF head $\phi_T$ predicts the needle tip position $P$, LLM $\phi_L$ generates action $\mathcal{A}=\phi_L(\epsilon_V(O_t), I, P)$, where $\epsilon_V(O_t)$ is the encoded vision embeddings.
$P$ is given by the bounding box $[x1,y1,x2,y2]$ of the needle tip, represented by its top-left $(x1,y1)$ and bottom-right $(x2,y2)$ coordinates in pixels.
The language instruction $I$ depicts the basic information and provides high-level command, given by 
\begin{quote}
    ``\textit{You are an ultrasound expert. The target position is [\texttt{TARGET}]. The insertion technique is [\texttt{TECH}]. Control the insertion based on visibility feedback: when visibility decreases, insert with a slower speed. Stop insertion upon reaching the target.}" 
\end{quote}
[\texttt{TARGET}] is the position of insertion target given by its center pixel coordinates.
[\texttt{TECH}] is the insertion technique (in-plane-static or in-plane-moving \cite{yan2023learning}).
The predicted action $\mathcal{A}$ is given by $\mathcal{A}=[\theta_{n}, v_{n}, \textbf{v}_{p}]$, where $\theta_{n}$ is the needle insertion angle, $v_{n}$ is the insertion speed, $\textbf{v}_{p}=[v_{p,x},v_{p,y},v_{p,z}]$ is the probe moving speed.
The needle and probe positions are then given by $x_n=v_n \Delta t + x_{n,0}$ and $\textbf{x}_p=\textbf{v}_p \cdot \Delta t + \textbf{x}_{p,0}$, where $x_{n,0}$ and $\textbf{x}_{p,0}$ are their initial positions.
When the needle reaches the target, the model generates a [\texttt{STOP}] token to terminate the procedure.

Leveraging the pretrained LLM, the proposed VLA-based control policy provides high-level reasoning based on global contextual information, in contrast to the ungeneralizable segmentation priors \cite{lapouge2020towards} and explicit needle deflection predictors \cite{khadem2016ultrasound} used in traditional needle steering controllers.
This language conditioning keeps the LLM architecture unchanged while encoding semantically rich control context.
Moreover, unlike traditional methods that require hyperparameter-controlled decision making, the proposed VLA model directly learns the control policy introduced by the expert from the training dataset.
The insertion can then be adaptively controlled by visual reasoning, thus avoiding lost tip tracking during low-visibility period and ensuring a lower safe velocity near complex tissue structures.

\subsection{2-Stage Training and Dataset Collection}
The proposed VLA model is trained by 2 stages.
Stage 1 is for pretraining on the tracking task, where only the CDF tracking head and TraCon register are trained and the remaining modules adopted from Qwen2.5-VL-3B \cite{bai2025qwen2} are frozen. 
A US needle tracking dataset $\mathcal{D}_1=\{(o_k,p_k)\}^N_{k=1}$ was collected for stage 1, where $o_k$ is the image, $p_k$ is the needle position ground truth measured by an optical tracker ClaroNav MicronTracker 3 (with an acceptable RMSE of 0.189 mm in our setup).
It was acquired using a Wisonic Clover 60 US machine and a Wisonic C5-1 convex transducer with an 18 gauge needle. $\mathcal{D}_1$ contains 41,075 frames from 239 videos ($1920\times 1080$) by 105 in-plane-static (IPS) trials and 134 in-plane-moving (IPM) trials \cite{kimbowa2024advancements,yan2023learning} at a velocity of 20 mm/s and 3 different insertion angles ($30^\circ$, $45^\circ$, $60^\circ$).
The needle and transducer (i.e. US probe) were manipulated by the aforementioned RUS system.
Several different materials were used for needle insertion, including phantoms made from fresh pork and solidified agar, as well as simulators composed of silicone and artificial tumors.
$\mathcal{D}_1$ is divided into training, validation, and testing sets in the ratio 7:1:2. \looseness=-1

Stage 2 is for fine-tuning the VLA model for adaptive needle insertion. Only the LLM and MLP aligner are fine-tuned with LoRA \cite{hu2022lora} while the remaining modules including CDF head and TraCon register are frozen.
The pretrained vision encoder is frozen during both stages.
A US needle insertion dataset $\mathcal{D}_2=\{(o_k,p_k,i_k,a_k)\}^N_{k=1}$ was collected using the same devices and setup as $\mathcal{D}_1$. In addition to image $o_k$ and needle positions $p_k$, $\mathcal{D}_2$ also includes language instructions $i_k$ and action ground truth $a_k$.
$a_k$ was collected through expert demonstrations, in which an experienced operator manually manipulated the needle to reach specified targets, following principles similar to the proposed uncertainty-based control policy. During these demonstrations, the needle velocity $v_n$, insertion angle $\theta_n$, and probe velocity $\textbf{v}_p$ were recorded in real time. The operator recorded a [\texttt{STOP}] sign upon reaching the target.
$\mathcal{D}_2$ contains 3,852 frames from 18 videos by 9 IPS trials and 9 IPM trials.
It is only used for training, not for testing.

\section{Experiments and Results}
\subsection{Implementation Details}
The evaluation was carried out on needle tracking and needle insertion respectively \footnote{See examples of needle tracking and insertion in supplementary video.}.
All experiments are implemented using PyTorch on a server with two NVIDIA A800 GPUs (although often considered high-end, A800 is based on an old NVIDIA Ampere architecture that is slower than current flagships).
For needle tracking evaluation, several state-of-the-art trackers are involved for comparison, including classic Siamese trackers \cite{li2019siamrpn++,guo2020siamcar,chen2022siamban}, transformer-based trackers \cite{lin2022swintrack,cui2022mixformer}, CNN-based trackers \cite{fu2021stmtrack}, and large-scale trackers \cite{lin2024tracking}.
All these models were trained on the same dataset (training set of $\mathcal{D}_1$) by the same strategy (350 epochs, batch size 48, AdamW optimizer) as ours.
Scaling, blur, and shifting were adopted for augmentation. The learning rate was set to 1e-4 and dropped by 10 after 100 epochs.

For needle insertion evaluation, the model pretrained on $\mathcal{D}_1$ was further fine-tuned on dataset $\mathcal{D}_2$ for a single epoch with a batch size of 16 and an AdamW optimizer. The learning rate was initialized to 1e-4 and modulated by a cosine annealing scheduler.

\subsection{Needle Tracking Evaluation and Ablation Studies}
\begin{figure}
    \centering
    \includegraphics[width=\linewidth]{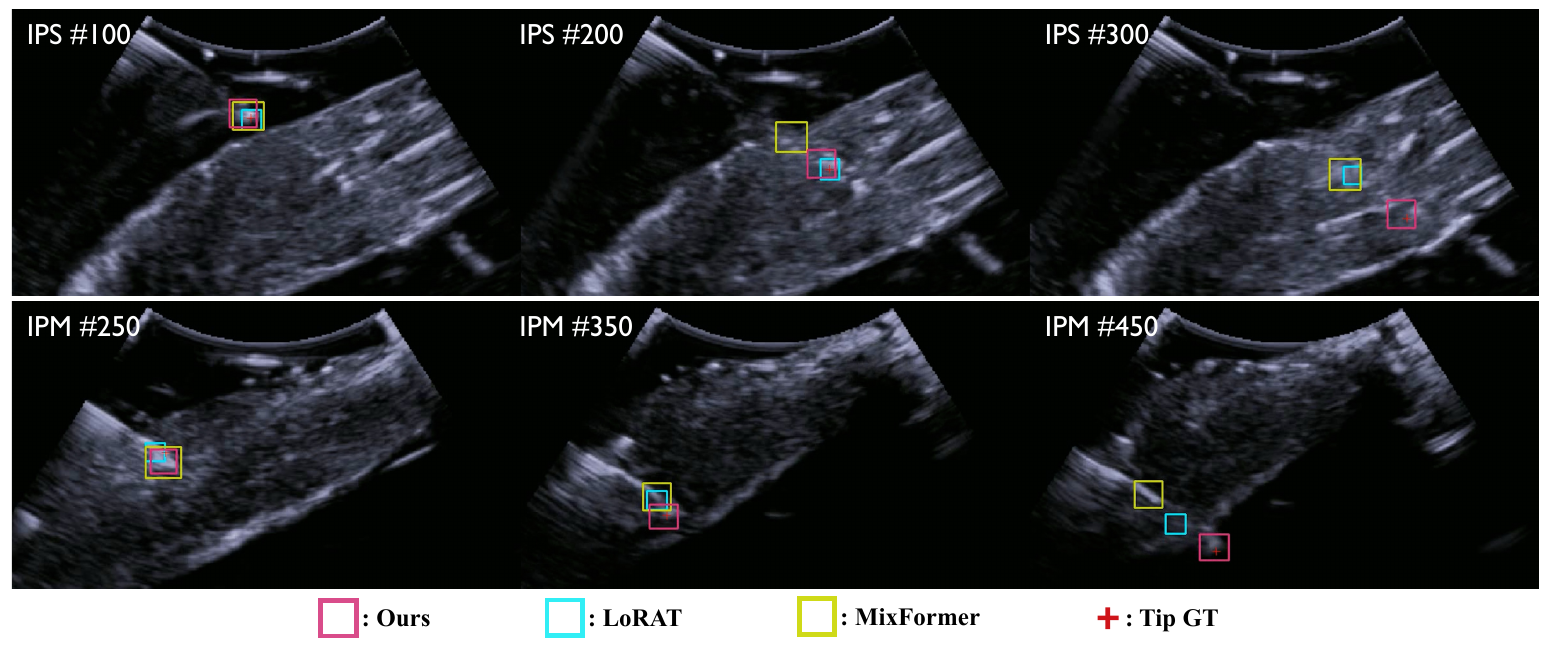}
    \caption{Needle tracking demonstration in a tissue phantom. The proposed method outperforms two state-of-the-art trackers under dynamic environment and degraded visibility. 
    }
    \label{fig_tracking}
\end{figure}

Models were trained and evaluated on the training set and testing set of $\mathcal{D}_1$. 
The result is reported as area under curve (AUC) \cite{wu2013online} and precision (\textit{P}) \cite{muller2018trackingnet} in percentage, as well as average error (Err) and standard deviation (SD) in mm.
As shown in Tab.~\ref{tab_tracking}, the proposed tracking pipeline achieves the best performance in almost all comparisons against the other SOTA trackers.
Our tracker achieves 10.7\% and 16.0\% improvement on Err and SD over the second best method, showing advancement on tracking accuracy and robustness.
Furthermore, our framework is the only approach that achieves a SD of less than 2 on both the IPS and IPM tasks.
The tracking demonstration in Fig.~\ref{fig_tracking} further illustrates that the proposed method attains the most robust and accurate tracking under challenging conditions.
This accurate tracking serves as the basis of successful automated insertion.

Ablation studies were performed on four variants, as shown in Tab.~\ref{tab_tracking}.
$\mathcal{V}_{T,1}$ and $\mathcal{V}_{T,2}$ investigate the TraCon register and the impact of its length $L_r$. The results in $\mathcal{V}_{T,1}$ indicate that using a longer register does not consistently improve performance and may even lead to undesired outcomes. In contrast, removing the register results in performance degradation across almost all metrics in $\mathcal{V}_{T,2}$. These findings demonstrate that the proposed TraCon register can effectively enhance tracking through model conditioning.
$\mathcal{V}_{T,3}$ and $\mathcal{V}_{T,4}$ investigate the significance of fusing cross-depth features. They represent variants where cross-depth semantic fusion is removed and only shallow feature or deep feature is kept respectively. The results show that receiving only shallow or deep features leads to degraded performance. This indicates that taking into account both shallow positional features and deep semantic features is critical for tracking.

\begin{table}
\caption{
Evaluation and ablation study results of needle tracking in AUC (\%), \textit{P} (\%), Err (mm), and SD (mm). The methods with the best and the second best performance are in {\color{r}red} and {\color{b}cyan}. 
}
\label{tab_tracking}

\begin{subtable}[b]{0.5\textwidth}
\centering
\setlength{\tabcolsep}{0.50mm}{
\scalebox{0.82}{
\begin{tabular}{l|ccc|ccc|ccc}
\toprule
\multirow{2}{*}{Method} & \multicolumn{3}{c|}{In-plane-static (IPS)}  & \multicolumn{3}{c|}{In-plane-moving (IPM)} & \multicolumn{3}{c}{Mean} \\
\cmidrule(lr){2-10}
& \cellcolor{c2!50}AUC$_\uparrow$ & \cellcolor{c2!50}$\textit{P}$$_\uparrow$ & \cellcolor{c2!50}Err$\pm$SD$_\downarrow$ & \cellcolor{c2!50}AUC$_\uparrow$ & \cellcolor{c2!50}$\textit{P}$$_\uparrow$ & \cellcolor{c2!50}Err$\pm$SD$_\downarrow$ & \cellcolor{c2!50}AUC$_\uparrow$ &  \cellcolor{c2!50}$\textit{P}$$_\uparrow$ & \cellcolor{c2!50}Err$\pm$SD$_\downarrow$ \\
\midrule
SiamRPN++ \cite{li2019siamrpn++} & 45.0 & 60.1 & 5.55$\pm$3.19 & 60.2 & 76.3 & 4.10$\pm$3.64 & 51.4 & 66.9 & 4.90$\pm$3.41 \\ 
SiamCAR \cite{guo2020siamcar} & 53.1 & 71.0 & 4.87$\pm$3.28 & 60.4 & 83.3 & 3.38$\pm$3.09 & 56.2 & 76.2 & 4.27$\pm$3.21 \\
SiamBAN \cite{chen2022siamban} & 53.0 & 72.7 & 4.75$\pm$3.30 & 65.4 & 92.0 & 4.13$\pm$2.99 & 58.2 & 80.8 & 4.50$\pm$3.19 \\
SwinTrack \cite{lin2022swintrack} & 49.6 & 70.6 & 4.99$\pm$3.37 & 65.8 & \color{b}94.0 & 3.41$\pm$2.93 & 56.4 & 80.4 & 4.37$\pm$3.09 \\ 
STMTrack \cite{fu2021stmtrack} & 52.1 & 74.1 & 4.96$\pm$3.20 & 64.3 & 92.5 & 3.17$\pm$3.03 & 57.2 & 81.9 & 4.16$\pm$2.98 \\
MixFormer \cite{cui2022mixformer} & 57.9 & 77.0 & 4.32$\pm$2.77 & \color{r}68.0 & 93.9 & 3.01$\pm$2.73 & \color{b}62.1 & 84.1 & 3.79$\pm$2.77 \\
LoRAT \cite{lin2024tracking} & \color{b}59.7 & \color{b}79.0 & {\color{b}3.82}$\pm${\color{b}2.55} & 65.1 & 93.7 & {\color{b}2.73}$\pm${\color{b}2.68} & 62.0 & \color{b}85.2 & {\color{b}3.37}$\pm${\color{b}2.62} \\
Ours & \color{r}60.3 & \color{r}84.0 & {\color{r}3.45}$\pm${\color{r}1.92} & \color{b}67.9 & \color{r}95.5 & {\color{r}2.29}$\pm${\color{r}1.98} & \color{r}63.5 & \color{r}88.9 & {\color{r}3.01}$\pm${\color{r}2.20}  \\
\midrule
\end{tabular}
}
}
\end{subtable}

\begin{subtable}[b]{0.5\textwidth}
\centering
\setlength{\tabcolsep}{1.0mm}{
\scalebox{0.92}{
\begin{tabular}{l|ccc}
\midrule
\multirow{2}{*}{Ablations} & \multicolumn{3}{c}{Mean} \\
\cmidrule(lr){2-4}
& \cellcolor{c2!50}AUC$_\uparrow$ & \cellcolor{c2!50}$\textit{P}$$_\uparrow$ & \cellcolor{c2!50}Err$\pm$SD$_\downarrow$ \\
\midrule
Baseline: default & 63.5 & 88.9 & 3.01$\pm$2.20 \\
$\mathcal{V}_{T,1}$: $L_r=16$ & 63.7{\scriptsize(+0.2)} & 88.5{\scriptsize(-0.4)} & 3.10{\scriptsize(+0.09)}$\pm$2.19{\scriptsize(-0.01)} \\
$\mathcal{V}_{T,2}$: $L_r=0$ (w/o TraCon R) & 62.8{\scriptsize(-0.7)} & 86.1{\scriptsize(-2.8)} & 3.18{\scriptsize(+0.17)}$\pm$2.10{\scriptsize(-0.10)} \\
$\mathcal{V}_{T,3}$: w/o fusion, only shallow & 62.1{\scriptsize(-1.4)} & 86.4{\scriptsize(-2.5)} & 3.49{\scriptsize(+0.48)}$\pm$2.62{\scriptsize(+0.42)} \\
$\mathcal{V}_{T,4}$: w/o fusion, only deep & 61.7{\scriptsize(-1.8)} & 87.0{\scriptsize(-1.9)} & 3.32{\scriptsize(+0.31)}$\pm$2.95{\scriptsize(+0.75)} \\
\bottomrule
\end{tabular}
}
}
\end{subtable}

\end{table}

\subsection{Needle Insertion Evaluation and Ablation Studies}
The model trained on $\mathcal{D}_1$ was further fine-tuned on dataset $\mathcal{D}_2$, where only LLM and MLP aligner were trained with LoRA.
The comparison is performed between the proposed VLA-based RUS insertion and manual insertion regarding success rate (SUC) in percentage and procedure time (T) in seconds.
For SUC, a successful attempt is defined as the needle tip reaching within 5 mm of the target point. Any deviation of the needle from the target or complete loss of needle visualization during the process is considered a failure. $T$ denotes the average completion time across all successful trials.
For VLA-based RUS insertion, a total of 40 attempts were performed (20 IPS and 20 IPM). For manual insertion, five experienced users each performed 4 IPS and 4 IPM insertions, also resulting in a total of 40 attempts (20 IPS and 20 IPM). 
Ablation studies on RUS insertion were conducted using the same protocol.
For each target position, both RUS and manual insertion were conducted.

This study does not compare with conventional hand-crafted pipelines or existing VLA models for US needle insertion, as such methods are scarce and lack open-source implementations. Instead, we focus on comparisons with manual operation, which remains the mainstream approach in clinical practice and directly reflects the practical benefits and translational potential for real-world adoption.
As shown in Tab.~\ref{tab_insertion}, the proposed VLA-based tracking-insertion pipeline achieves a significant 33.3\% SUC improvement with less time consumption than manual insertion. For IPM insertion, the SUC improvement even reached 63.6\% with an average time reduction of 7.1 s.
Furthermore, with the proposed asynchronous pipeline, the average frame rates for tracking and action generation reached 25.1 FPS and 10.4 FPS, respectively.
It can not only provide operators with real-time needle position feedback, but also ensure a safe and acceptable action generation frequency, which surpasses that of most existing VLA models \cite{ng2025endovla,li2024robonurse}.

\begin{figure}
    \centering
    \includegraphics[width=\linewidth]{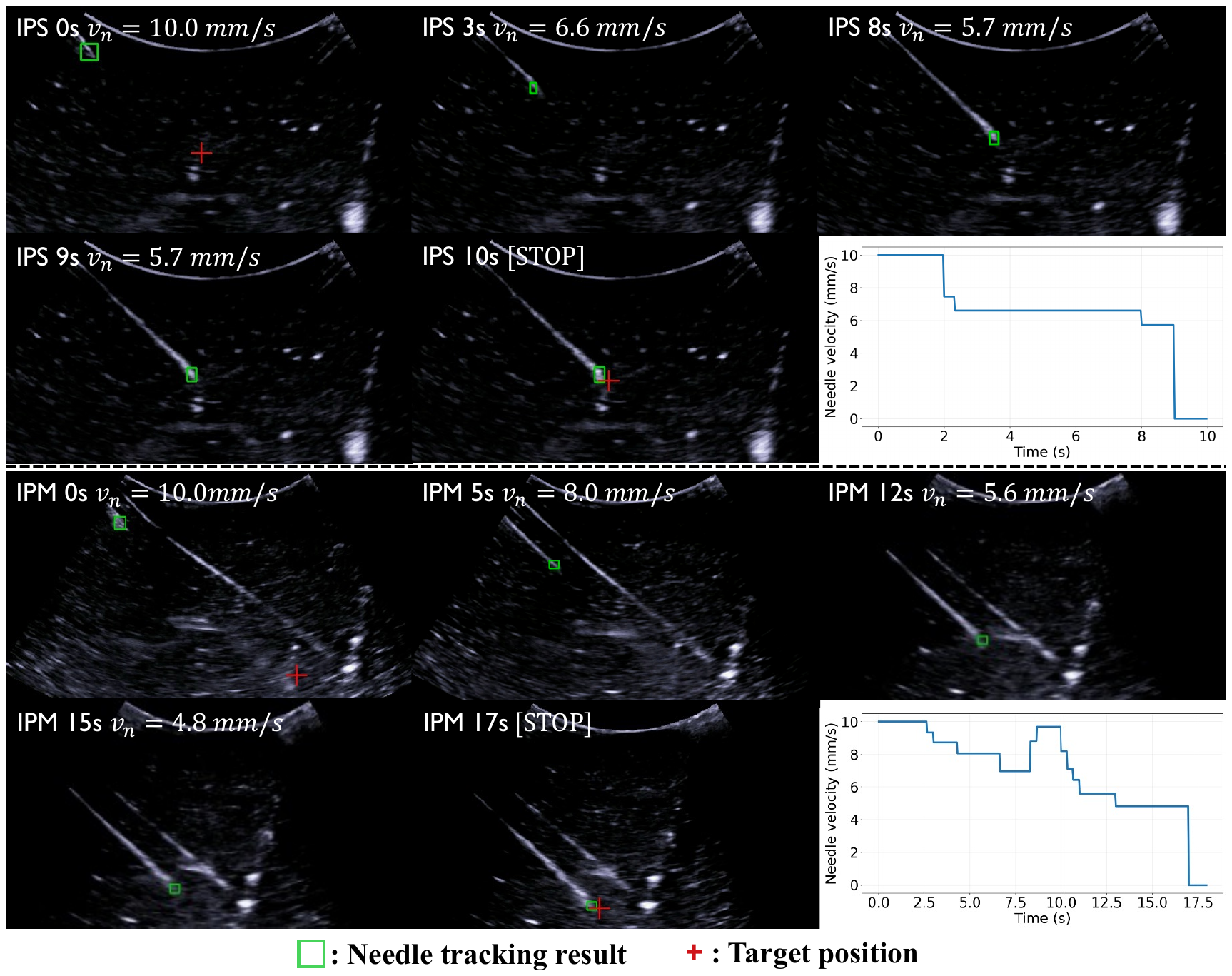}
    \caption{Two examples of adaptive needle insertion with uncertainty-aware control. The plot of $v_n$ and $t$ is given. The target position is specified by the operator. The proposed framework proactively responds to environmental uncertainty by adjusting the speed when the needle is occluded by tissue to ensure tracking and safety.}
    \label{fig_insertion}
    \vspace{-3mm}
\end{figure}

Two examples are shown in Fig.~\ref{fig_insertion}, where the insertion speed is dynamically adjusted by the proposed VLA framework to ensure consistent needle visualization and improved outcomes.
In the IPM case, needle insertion is slowed during the initial stage due to increased uncertainty. The speed gradually recovers in the middle stage, and as the needle approaches the target, insertion slows again to account for ambiguity caused by tissue occlusion. This dynamic adaptation allows the needle to respond appropriately at any position, thereby improving the success rate.

Ablation studies were performed on four variants.
$\mathcal{V}_{I,1}$ adopts the same structure as $\mathcal{V}_{T,1}$ in Tab.~\ref{tab_tracking}. The performance degradation demonstrates that poor tracking leads to a higher rate of insertion failures.
Without the TraCon register, $\mathcal{V}_{I,2}$ results in more insertion failures and increased time consumption.
It further demonstrates the significance of PEFT enabled by the TraCon register, highlighting its substantial improvement of PEFT results even in scenarios with extremely limited training data (only 18 videos in $\mathcal{D}_2$).
$\mathcal{V}_{I,3}$ shows a significant SUC decrease when the asynchronous pipeline is removed, due to the inability to respond promptly to the latest frames. Since the entire pipeline operates synchronously, tracking and action generation must share an inference speed of 13.1 FPS, which falls short of real-time requirements and results in an inevitable discrepancy between the estimated and actual needle positions.
$\mathcal{V}_{I,4}$ removes the uncertainty-aware control policy, inserting the needle at a constant velocity of $10~\mathrm{mm/s}$ and stopping upon reaching the target. The observed decrease in SUC indicates that, when uncertainty increases, continuing insertion at a fixed speed without intervention can lead to tracking loss and insertion failure. This highlights the importance of controlling insertion with environment awareness.

\begin{table}
\caption{
Evaluation and ablation study results of needle insertion in SUC (\%), T (s). 
}
\label{tab_insertion}

\begin{subtable}[b]{0.5\textwidth}
\centering
\setlength{\tabcolsep}{2.17mm}{
\scalebox{0.88}{
\begin{tabular}{l|cc|cc|cc}
\toprule
\multirow{2}{*}{Method} & \multicolumn{2}{c|}{\scriptsize In-plane-static (IPS)}  & \multicolumn{2}{c|}{\scriptsize In-plane-moving (IPM)}  & \multicolumn{2}{c}{Mean} \\
\cmidrule(lr){2-7}
& \cellcolor{c2!50}SUC$_\uparrow$ (\%) & \cellcolor{c2!50}T (s) & \cellcolor{c2!50}SUC$_\uparrow$ (\%) & \cellcolor{c2!50}T (s) & \cellcolor{c2!50}SUC$_\uparrow$ (\%) & \cellcolor{c2!50}T (s) \\
\midrule
Ours & \textbf{70.0} & 12.1 & \textbf{90.0} & 23.9 & \textbf{80.0} & 17.3 \\ 
Manual & 65.0 & 17.1 & 55.0 & 31.0 & 60.0 & 23.2 \\ 
\midrule
\end{tabular}
}
}
\end{subtable}

\begin{subtable}[b]{0.5\textwidth}
\centering
\setlength{\tabcolsep}{4.00mm}{
\scalebox{1.0}{
\begin{tabular}{l|cc}
\midrule
\multirow{2}{*}{Ablations} & \multicolumn{2}{c}{Mean} \\
\cmidrule(lr){2-3}
& \cellcolor{c2!50}SUC$_\uparrow$ (\%) & \cellcolor{c2!50}T (s) \\
\midrule
Baseline: default & 80.0 & 17.3 \\  
$\mathcal{V}_{I,1}$: $L_r=16$ & 75.0{\scriptsize(-5.0)} & 16.1{\scriptsize(-1.2)} \\
$\mathcal{V}_{I,2}$: $L_r=0$ (w/o TraCon R) & 72.5{\scriptsize(-7.5)} & 24.5{\scriptsize(+7.2)} \\
$\mathcal{V}_{I,3}$: w/o \textit{async} pipeline & 67.5{\scriptsize(-12.5)} & 15.0{\scriptsize(-2.3)} \\ 
$\mathcal{V}_{I,4}$: w/o \textit{uncertainty} control & 70.0{\scriptsize(-10.0)} & 8.7{\scriptsize(-8.6)} \\ 
\bottomrule
\end{tabular}
}
}
\end{subtable}
\vspace{-4mm}
\end{table}

\subsection{Discussion}
The improvement can be attributed to two factors: first, the \textit{efficient adaptation of the large-scale pretrained vision backbone}; second, the \textit{LLM-enabled high-level reasoning}.

In contrast to existing large-scale vision trackers like LoRAT \cite{lin2024tracking} that directly use the deepest layer outputs of large backbones, the proposed CDF tracking head in this work efficiently fuses and utilizes multi-level features from different layers of the deep vision encoder, avoiding positional information loss like LoRAT.
It can be observed in Tab.~\ref{tab_tracking} that our method outperforms LoRAT on most metrics, benefiting from the CDF head’s cross-depth fusion. 
Besides, unlike LoRAT, this work does not fine-tune the vision encoder for tracking, as fine-tuning the vision encoder shared by both the tracking head and the LLM could introduce feature imbalance between them, which would negatively affect LLM-based action generation.
Furthermore, ablation studies on the TraCon register show that the vision encoder can be \textbf{externally conditioned} instead of internally fine-tuned, even when utilizing a minimal number of trainable parameters.
Such a lightweight PEFT approach is crucial for preventing model overfitting in US tasks, given that US datasets usually consist of relatively few samples and insufficient diversity.
These improvements ensure \textit{accurate needle position tracking} even in dynamic environments.

Based on accurate needle position feedback, adaptive needle insertion control can be achieved. 
The results in Tab.~\ref{tab_insertion} demonstrate the effectiveness of the proposed framework. The LLM is trained on large-scale open-world data, inherently capturing both semantic information and physical commonsense. Compared to the subjective and user-dependent nature of manual insertion, the proposed method leverages the LLM’s \textit{generalizable environment-aware capability} to enable adaptive control. When imaging is affected by occlusion, artifacts, or intermittent needle visibility, the proposed control policy can make context-aware decisions to ensure continuous visualization of the needle.

Last but not least, the asynchronous VLA pipeline enables non-interfering synergy between tracking and control, ensuring real-time tracking and action generation. 
This contributes to a higher insertion success rate, as shown in Tab.~\ref{tab_insertion} ($\mathcal{V}_{I,3}$). As such, the asynchronous pipeline is an indispensable component of the proposed VLA insertion-tracking model.

\section{CONCLUSIONS}
In this paper, a VLA framework is proposed for adaptive US-guided needle insertion and tracking. Extensive experiments demonstrate that our framework achieves state-of-the-art tracking accuracy and significantly improves insertion success rates compared to manual operation. These results highlight the potential of VLA models for enhancing safety and efficiency in RUS needle insertion.
One limitation is that the tracking speed only barely meets real-time requirements, indicating room for further efficiency improvements. In future work, multi-degree-of-freedom probe manipulation will be developed to proactively enhance needle visibility.
A dataset with a larger cohort will be collected in the future.






\bibliographystyle{ieeetr}
\bibliography{reference}

\end{document}